\documentclass{article}

\usepackage[english]{babel}

\usepackage{authblk}
\usepackage{cite}
\usepackage{amsmath,amssymb,amsfonts}
\usepackage{algorithmic}
\usepackage{graphicx}
\usepackage{textcomp}
\usepackage{xcolor}
\usepackage{footnote}

\usepackage{graphicx}
\usepackage{amssymb}
\usepackage{booktabs}
\usepackage{multirow}
\usepackage{color}
\usepackage{algorithm}
\newcommand{\ppm}{\,$\pm$\,}

\newcommand{\modif}[1]{#1}

\def\BibTeX{{\rm B\kern-.05em{\sc i\kern-.025em b}\kern-.08em
    T\kern-.1667em\lower.7ex\hbox{E}\kern-.125emX}}
\begin{document}

\title{CE-Dedup: Cost-Effective Convolutional Neural Nets Training based on Image Deduplication}

\author[1]{Xuan Li}
\author[1]{Liqiong Chang}
\author[1]{Xue Liu}
\affil[1]{School of Computer Science\\
McGill University\\
Montreal, Quebec, Canada}

\maketitle

\begin{abstract}
Attributed to the ever-increasing large image datasets, Convolutional Neural Networks (CNNs) have become popular for vision-based tasks. It is generally admirable to have larger-sized datasets for higher network training accuracies. However, the impact of dataset quality has not to be involved. It is reasonable to assume the near-duplicate images exist in the datasets. For instance, \modif{the Street View House Numbers (SVHN) dataset having cropped house plate digits from 0 to 9 are likely to have repetitive digits from the same/similar house plates. } 
Redundant images may take up a certain portion of the dataset without consciousness. While contributing little to no accuracy improvement for the CNNs training, these duplicated images unnecessarily pose extra resource and computation consumption. 
To this end, this paper proposes a framework to assess the impact of the near-duplicate images on CNN training performance, called CE-Dedup. Specifically, CE-Dedup associates a hashing-based image deduplication approach with downstream CNNs-based image classification tasks. 
CE-Dedup balances the tradeoff between a large deduplication ratio and a stable accuracy by adjusting the deduplication threshold. The effectiveness of CE-Dedup is validated through extensive experiments on well-known CNN benchmarks. 
On one hand, \modif{while maintaining the same validation accuracy, CE-Dedup can reduce the dataset size by 23\%. On the other hand, when allowing a small validation accuracy drop (by 5\%), CE-Dedup can trim the dataset size by 75\%}.

\end{abstract}


\section{Introduction}\label{sec:introduction}
Empowered by the large volume of data, deep neural networks have achieved impressive results in natural language processing, computer vision, and reinforcement learning. In particular, Convolutional Neural Networks (CNNs) achieve near human-level performance on a variety of vision-based tasks. This success is primarily attributed to the enormous volume of image data. Well-known datasets such as CIFAR100~\cite{cifar100}, ImageNet~\cite{deng2009imagenet}, COCO~\cite{lin2014microsoft} have images ranging from several thousand to millions.

In general, with a larger volume of datasets, the trained network has higher accuracy and more robustness. However, the quality of the datasets has not been assessed~\cite{li2020intelligent}. There is inevitably redundant information in the million-scale datasets. 
For example, the SVHN dataset contains cropped digits from street view house number plates. It is reasonable that exists repetitive images with similar plate materials and the same printed digits. These redundant images (we refer to as near-duplicates) bring two negative impacts.

\textbf{Excessive resource consumption}. 
Storing these redundant images would consume huge resources and costs. For instance, ImageNet~\cite{deng2009imagenet} holds over 1.2 million images, taking up 150 GB of disk space. This situation is aggravated when training is distributed, as each cluster needs to hold a copy of the dataset. To train a moderate CNN, ResNet50~\cite{he2016deep} in parallel, ~\cite{goyal2017accurate} uses a server of over 250 clusters, which amounts to 37.5 TB data in total. Training on these datasets will also impose excessive resources and costs.  Microsoft and OpenAI~\cite{scott_2020} have worked together to develop a supercomputer dedicated to the GPT-3 model training~\cite{brown2020language}. The supercomputer consists of over 285,000 CPUs, 10,000 GPUs, costing \$4.6 million only for the training process, not to mention the carbon emissions caused by the training and wireless data transmission~\cite{su2012secure}.

\begin{figure*}[t!]
\centering
\setlength{\tabcolsep}{3pt} 
\begin{tabular}{cc}
\includegraphics[width=.18\textwidth]{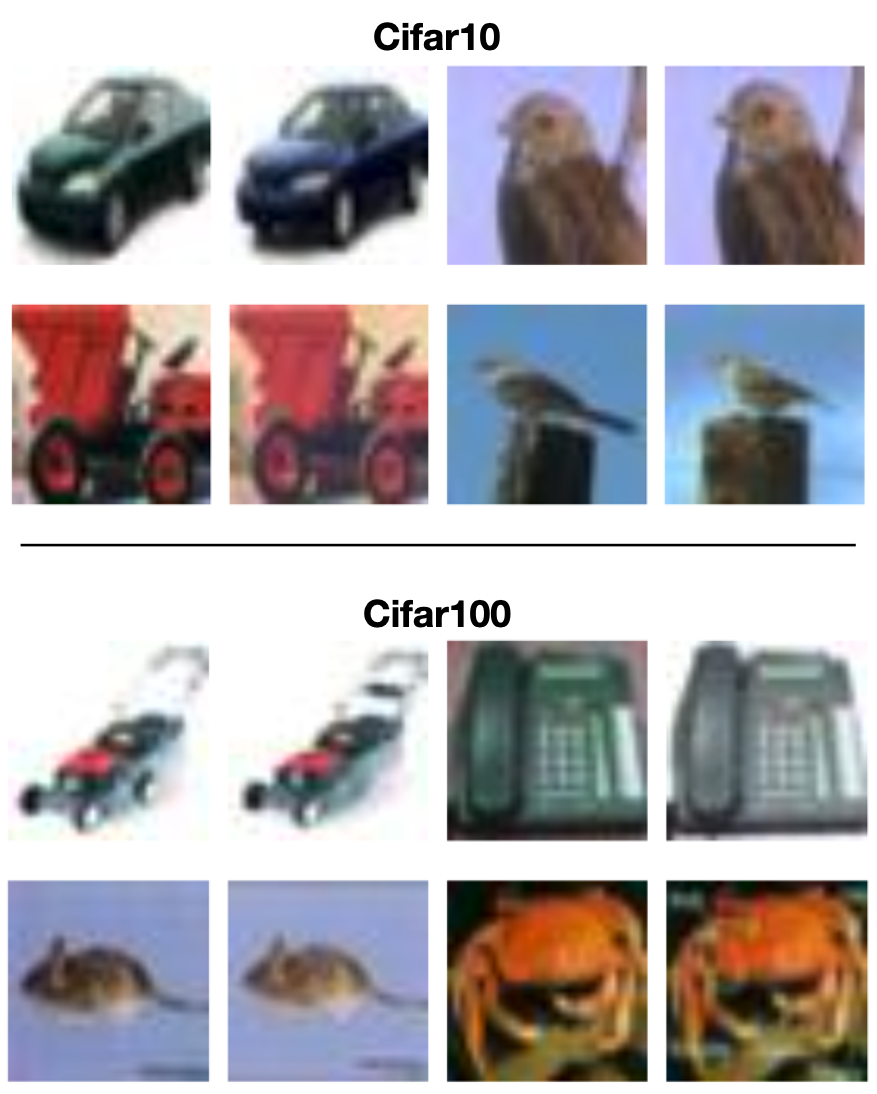} &
\includegraphics[width=.7\textwidth]{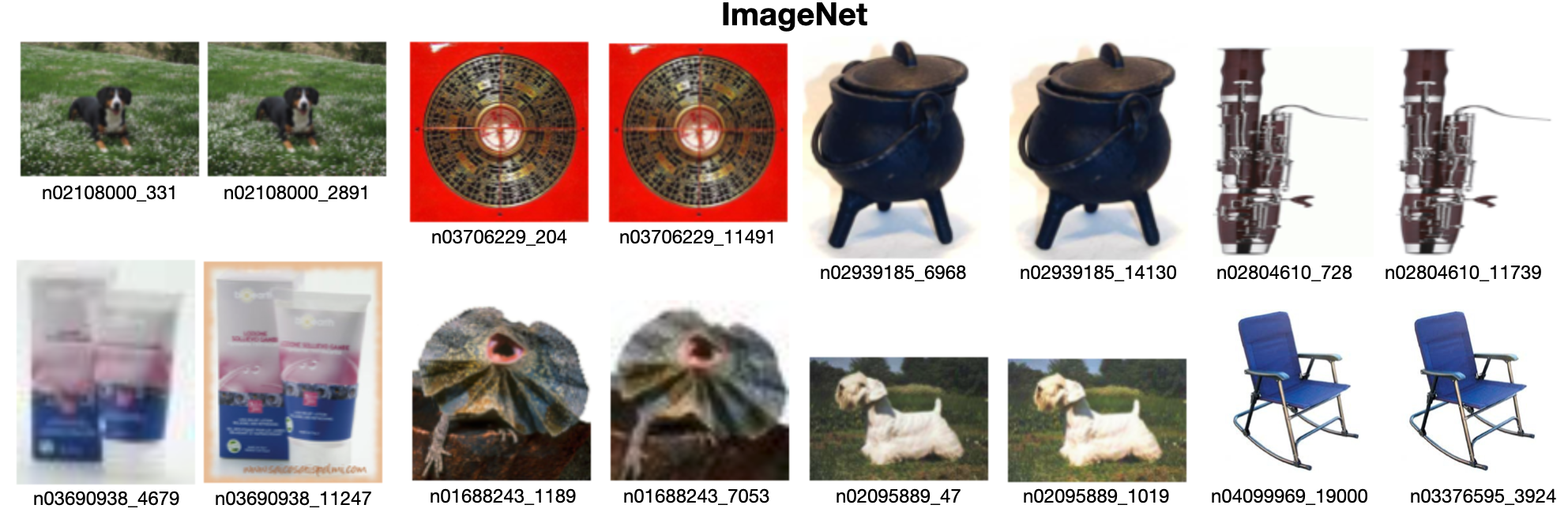} \\
(a) & (b) \\

\end{tabular}
\caption{Sampled ND image pairs from the CIFAR(a) and ImageNet(b) datasets. In each pair, the images have high visual similarity where each image can be derived with certain transformation from each other. \modif{Original file names associated with each image are only shown for ImageNet due to the reason that Cifar10 and Cifar100 do not provide the individual file names.}}
\label{fig:nd_images}
\end{figure*}

\textbf{Little to no accuracy and robustness improvement}. 
CNNs learn little information from these redundant images. For example, Fig.~\ref{fig:nd_images}(a) shows 8 pairs of near-duplicate (ND) images existing in the CIFAR10 and CIFAR100 datasets. The images in each pair can be derived from each other by either color or geometric transformations, \modif{which has been the de facto standard as data augmentation approaches in pre-processing}. Thus, \modif{storing and training} those ND images would have limited contribution to the accuracy and robustness improvement. What's worse, the ND images may be split into training and testing sets without awareness. We observe that 6 out of the 8 ND image pairs in Fig.~\ref{fig:nd_images}(a) are separated in the training and testing sets \modif{(the training and testing sets are made officially from ~\cite{pytorch2021})}. \modif{Fig.~\ref{fig:nd_images}(b) shows samples of ND images with the file names from the original ImageNet~\cite{deng2009imagenet}}. As a result, the trained CNNs achieve high accuracy on the testing set. However, they will suffer drastic accuracy drops and even fail in practical deployment. Such issues are hard to detect in real-life applications.



It is therefore essential to review the huge amount of data and remove the redundant images. Image deduplication is a popular technique widely used by cloud service providers to recognize and remove duplicate images. To find duplicates, many hashing-based image deduplication algorithms are proposed. The main concept is to encode images into hashes, then find the duplicates by calculating the hamming distances between all pairs of hashes. If a distance is less than a pre-defined threshold, the two images are considered duplicates. Thus, one copy can be discarded. After removing the duplicated images, the storage size for the dataset is significantly reduced while the discriminant features in the images are still kept. 


\modif{We consider incorporating image deduplication into a CNN-training pipeline, which to our knowledge, is the first of proposing such strategy. Deduplication performance is controlled by a pre-defined threshold comparing the pair-wise hamming distances. In general, a small threshold} brings a slightly decreased dataset. Hypothetically, CNNs trained on such a dataset should have test accuracies comparable to those trained on the original dataset. While a \modif{large threshold} enables a greatly size-reduced dataset, but most likely would result in degraded test accuracy. Ideally, one would like to find the \modif{optimal threshold} that reduces the dataset volume while also maintains a high test accuracy.



Towards this end, we propose \textit{CE-Dedup}, a framework that combines hashing-based image deduplication algorithms with CNN classification tasks to achieve cost-effective CNN training. 


In summary, we make the following contributions:

\begin{itemize}
  \item We propose \textit{CE-Dedup} \modif{which, to our knowledge, is the first approach that combines near-duplicate images reduction with image classification tasks. It incorporates a hashing-based image deduplication approach with CNN training. By controlling the distance threshold, one can balance between the largest data size reduction and the highest accuracy retain.} The framework \textit{CE-Dedup} will be made publicly available.
  
  \item We analyze and characterize the performance impact brought by hashing-based image deduplication under different setups. Through comprehensive experiments on large-scale datasets, we show the SVHN dataset can be reduced by over 23\% while maintaining the same accuracy as the original one. When the SVHN is deduplicated by over 75\%, we observe a 5\% accuracy drop.
\end{itemize}


\begin{figure*}[ht!]
\centering
\setlength{\tabcolsep}{5pt} 
\begin{tabular}{c}
\includegraphics[width=.85\textwidth]{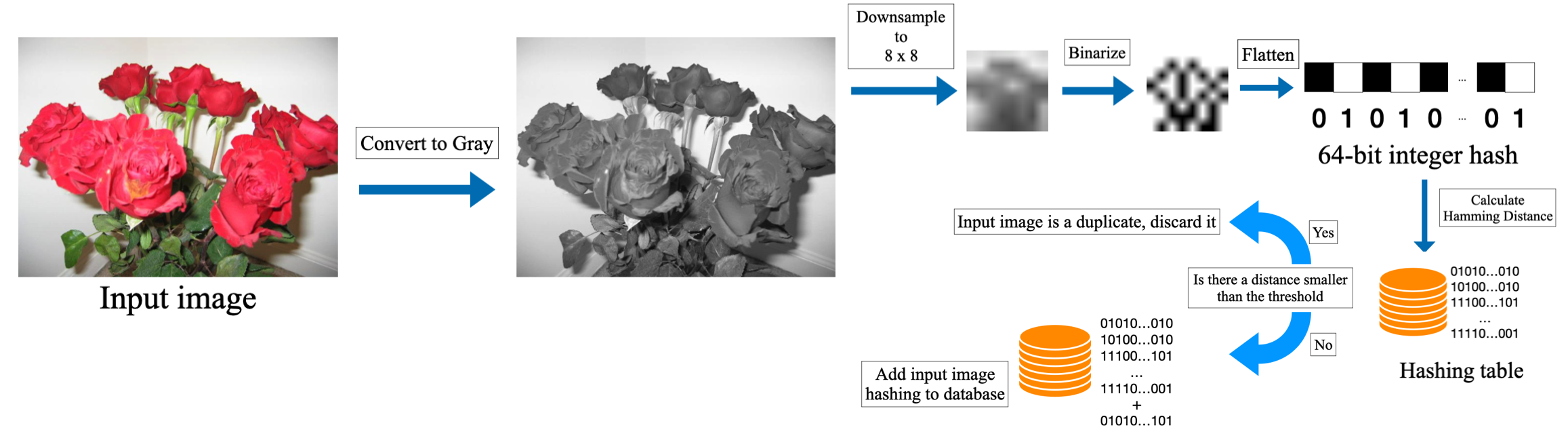} \\
\end{tabular}
\caption{A general workflow of hashing-based image deduplication. Each image is processed and converted to a 64-bit integer hash. The hash value is compared with existing hash values using hamming distance. If there exists a distance smaller than the threshold, the input is considered as a duplicate. Otherwise, the image is unique, and the image hash is added to the hashing table.}
\label{fig:hash_workflow}
\end{figure*}

\section{Framework Overview}
The Algorithm~\ref{ref:algo} shows the pipeline of \textit{CE-Dedup}. It consists of two stages, image deduplication, and CNN training. 

In stage 1, we apply various advanced hashing-based image deduplication approaches on several commonly used large-scale datasets. With different threshold settings, we assess the deduplication efficacy of each approach from aspects of 1) hashing time, 2) memory consumption, and 3) average duplicates found per image.

In stage 2, we train CNNs on deduplicated datasets. The 1) validation accuracy, and 2) training time derived from the original datasets and the deduplicated datasets are compared to quantify the effectiveness of \textit{CE-Dedup}. We also explore the duplicates that exist in these large-scale datasets and how would deduplication help boost the performance of CNNs.

In the following two sections (Section \ref{sec:hashing} and \ref{sec:cnn}), the details of each stage will be presented respectively. Then Section \ref{sec:experiment} introduces the datasets used and CNN training setups in extensive experiments. In Section \ref{sec:result}, the performances of deduplication and CNN training are evaluated. Section \ref{sec:related} reviews related works to hashing-based image deduplication techniques. Finally, Section \ref{sec:conclu} concludes the proposed framework with an in-depth discussion.

\begin{algorithm}[t]
\centering
\caption{CE-Dedup pipeline}\label{ref:algo}
\begin{algorithmic} 
\vspace{.5em}
\fontsize{8}{8}\selectfont
\STATE \emph{\textbf{Stage 1 -- Image Deduplication}:}
\vspace{.1em}
\STATE Convert images to grayscale;
\STATE Resize images to $8 \times 8$;
\STATE Binarize images based on a specified hashing algorithm;
\STATE Flatten each hash matrix into a 1D vector;
\STATE Calculate the Hamming distance $D_{i,j}$ for each hash vector pair $i, j$, and compare with a pre-defined threshold $th$;
\STATE \IF{$ D_{i,j} > th$}
\STATE $i, j$ are distinct images;
\ELSE
\STATE $i, j$ are ND images, discard one;
\ENDIF

\vspace{.5em}
\STATE \emph{\textbf{Stage 2 -- CNN training}:}
\vspace{.1em}
\STATE Perform data augmentation on the deduplicated dataset, including,
\begin{itemize}
\item random color perturbation;
\item random rotation;
\item random affine transformation;
\item random crop;
\end{itemize}
\STATE Train the CNN using cross-entropy loss;
\STATE Record the Validation accuracy;


\end{algorithmic}
\end{algorithm}

\section{Hashing-based Image Deduplication}\label{sec:hashing}

\subsection{Deduplication Workflow}
A general workflow for hashing-based image deduplication is shown in Fig~\ref{fig:hash_workflow}. It involves five steps as following,
\begin{itemize}
    \item \textbf{Pre-processing}. RGB Images are converted to grey scales, followed by down-sampling to reduce the resolution. Most algorithms resize images to $8 \times 8$ pixels.
    
    \item \textbf{Hashing}. Each pixel in the down-sampled images is converted from the original unsigned 8-bit integer ranging from 0 to 255, to a single binary value, either 0 or 1. This resulting $8 \times 8$ binary matrix is subsequently flattened into a 64-bit integer, which represents the original image. 
    
    \item \textbf{Storing}. Hash values are stored in a table for an efficient query. \modif{Since each image hash string takes up 113 bytes ram memory, million-scale dataset such as ImageNet costing over 100Mb can easily be handled by modern computer}. Commonly used data structures include KD-Tree, hierarchical K-means, and $k$-nearest neighbor.
    
    \item \textbf{Distance Calculating}. To compare the similarities between any two image hashes, the Hamming distance is used to compute the number of matched bits.
    
    \item \textbf{Deduplication}. Only the unique image is kept and duplicates are discarded \modif{(as shown in Fig~\ref{fig:nd_images}, which image is kept would not be significantly different for downstream classification task).  A threshold $th$ ($th \in [0,63]$) that allows for the maximum number of different bits in duplicates is pre-defined\footnote{We show in our experiment a plausible range would be $\in [0,12]$.}.} The value of 0 means that duplicates must have all bits to be the same. And the value of 12 indicates that duplicate images have less than or equal to 12 same hash bits.
\end{itemize}





\begin{figure*}[t!]
\centering
\setlength{\tabcolsep}{5pt} 
\begin{tabular}{c}
\includegraphics[trim=15 10 15 10,clip,width=.70\textwidth]{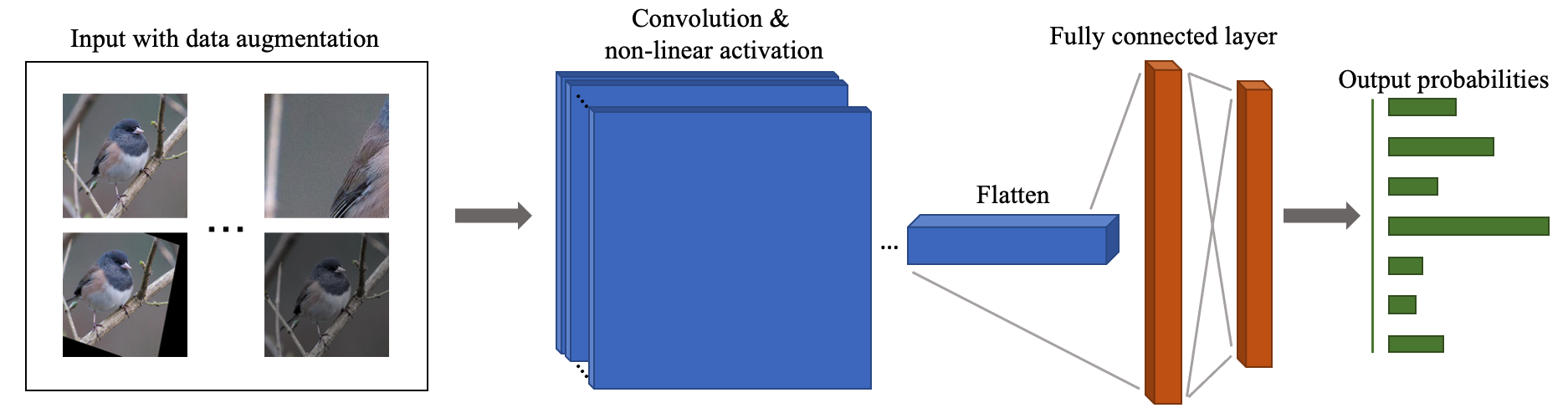}
\end{tabular}
\caption{A general workflow for CNN training. Input images are processed with data augmentation and then fed into the CNN to generate output probabilities.}
\label{fig:cnn_diagram}
\end{figure*}

\subsection{Hashing Algorithm Review}\label{subsec:hashing}
Image hashing analyzes image content and then creates a value based on its unique identification details. As shown in Fig~\ref{fig:hash_workflow}, a hash function is applied to the input image, and a hash value is computed based on the visual appearance of the image. In image deduplication, similar pictures should have similar hashes as well. In this way, ND images can be detected by simply compare hash differences.

This paper considers four widely used hashing techniques, AHash, DHash, PHash, and WHash. Their results for stage 1 including hashing time $Time_{Hash}$, duplicates found $Dups$, reduced dataset size $Size$, memory cost $Mem$, as well as results for stage 2 including CNNs training time on the deduplicated datasets $Time_{Train}$ and validation accuracy $Acc$ are presented. The hashing code is primarily based on~\cite{doi_ulrich_2021,idealods2019imagededup}.


\subsubsection{AHash} 
The Average Hashing (AHash) algorithm~\cite{chamoso2017hash} removes high frequencies and details by down-sampling image to pixels with size $N \times N$  ($8 \times 8$ in this paper). The RGB image is subsequently converted to grey scale and the average pixel value $P_{mean}$ is calculated. To convert each pixel $P_{val}$ to a bit value $P_{bit}$, $P_{val}$ is compared to the $P_{mean}$ and assigned 1 if greater than $P_{mean}$ otherwise 0. 

Then the $8 \times 8$ binary matrix is flattened into 64-bit integer to obtain the final hash. AHash is a simple and fast hashing algorithm. However, directly comparing with the mean may not yield stable results.

\subsubsection{DHash} 
The Difference Hashing (DHash) algorithm~\cite{chamoso2017hash} firstly resizes image with resolution of $N \times (N+1)$ ($8 \times 9$ in this paper). The RGB image is then converted to grey scales. To binarize the pixels, DHash compares the adjacent pixels. Specifically, each pixel is set to 1 if its pixel value is larger than the right pixel, otherwise 0.

This operation generates an $8 \times 8$ binary matrix. DHash algorithm has an improvement over AHash by comparing adjacent pixel pairs, thus more local patterns can be reserved.

\subsubsection{PHash}
The Perceptual Hashing (PHash) algorithm~\cite{zauner2010implementation} firstly resizes the image to pixels with a resolution of $32 \times 32$. Then, it applies Discrete Cosine Transform (DCT)~\cite{ahmed1974discrete} to turn spatial RGB values into a collection of frequencies and magnitudes. To remove the high-frequency components, only the top-left 2-dimensional matrix with size of $8 \times 8$ is kept. The mean value of the matrix is computed to generate the binary matrix, and further flattened to a 64-bit integer as the hash representation. PHash is extensively adopted due to its efficient DCT implementation, which is more robust to certain transformations.

\subsubsection{WHash}
The Wavelet Hashing (WHash) algorithm~\cite{singh2017robust} is similar to PHash, except that it uses Discrete Wavelet Transform (DWT)~\cite{shensa1992discrete} to convert spatial RGB values to frequencies and magnitudes. 



\begin{figure*}[t!]
\centering
\setlength{\tabcolsep}{3pt} 
\begin{tabular}{cc}
\includegraphics[width=.45\textwidth]{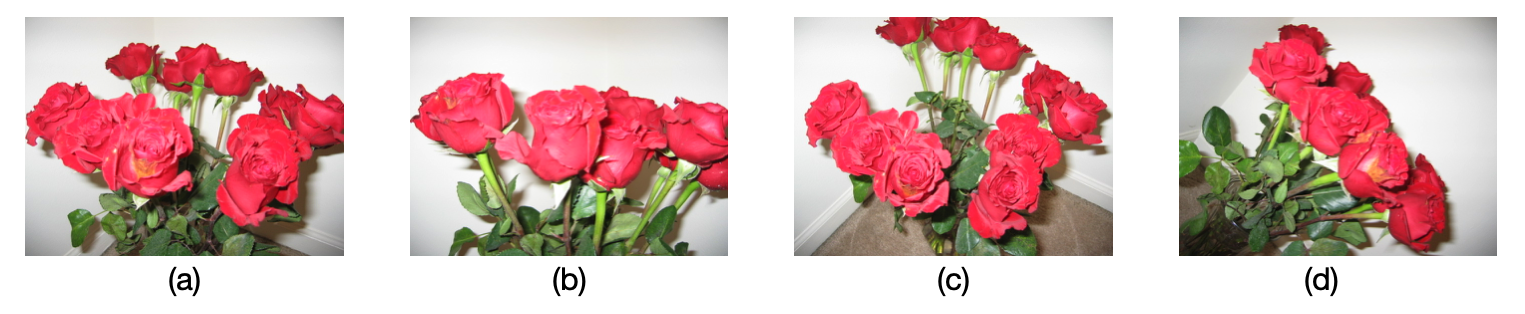} &
\includegraphics[width=.50\textwidth]{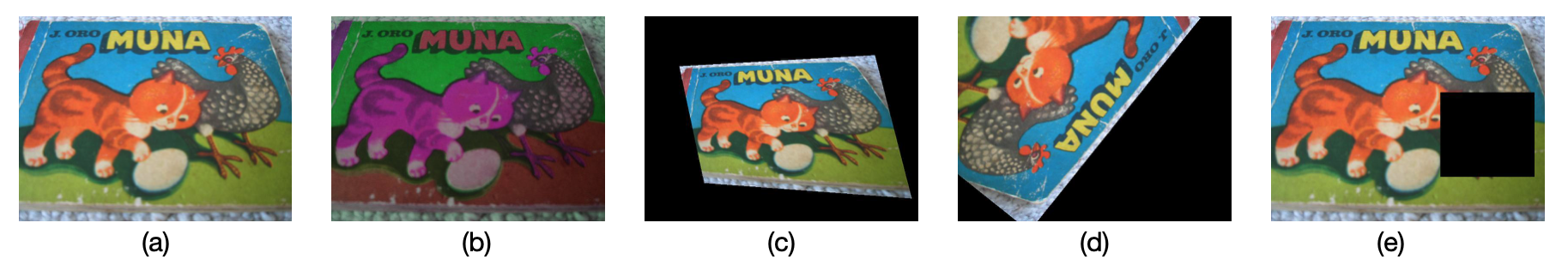} \\
(I) & (II) \\

\end{tabular}
\caption{\modif{two groups of ND images from the original UKBench dataset(I) and augmented UKBench(II) with color perturbation, perspective transformation, random rotated crop, and random erasing.}}
\label{fig:ukbench}
\end{figure*}

\section{CNN Training}\label{sec:cnn}
CNN is designed to process data with a rigid grid pattern, such as 1-D signal, 2-D images, and 3-D Voxels. It processes the spatial features from low to high-level hierarchical order. Fig~\ref{fig:cnn_diagram} introduces a general workflow of CNN training. Please refer to the paper ~\cite{o2015introduction} for more details.

A CNN takes an image as the input and processes it layer by layer through feature transformation to output a probability distribution. Cross-entropy loss is adopted to optimize the probability distribution toward the ground truth by back-propagating the loss to the parameters of the CNN.
For better generalizability, data augmentation techniques are commonly adopted to pre-process input images. Those techniques include random color perturbation, random rotation, random affine transformation, and random crop~\cite{bloice2017augmentor}.

\section{Experiments Setups}\label{sec:experiment}

\subsection{Datasets}
Five datasets are used including \modif{CIFAR10, CIFAR100, SVHN, UKBench, and Aug-UKBench.}

\begin{itemize}
    \item \textbf{CIFAR10}
    The CIFAR10 dataset contains a total of 60000 images from 10 classes of natural objects. Classes include the airplane, automobile, bird, cat, deer, dog, frog, horse, ship, and truck. It is a well-known benchmark dataset used by many classification algorithms. 
    
    \item \textbf{CIFAR100}
    The CIFAR100 dataset comprises 100 classes with 600 images in each class. Classes are a wide range of natural objects from animals such as dogs, whales to objects such as trains, motorcycles, etc.
    
    \item \textbf{SVHN}
    The Streat View and House Number (SVHN) dataset has cropped digits of images taken from house numbers. It has a total of 99287 images. Each image has a digit from 0 to 9 as the ground truth. 
    
    \item \textbf{UKBench}
    The UKBench dataset is a well-known benchmark used in image deduplication. In the dataset, every four ND images are different viewpoints taken from one object and organized as one group. There are a total of 2550 groups accounting for 10200 images in the dataset. Fig~\ref{fig:ukbench}(I) shows one group from the dataset.

    \item \textbf{AUG-UKBench}
     We augment the UKBench dataset and construct the AUG-UKBench dataset. One image of each group from UKBench is chosen, followed by the common data augmentation operations including random color perturbation, random perspective transformation, resized crop, and random erasing, respectively. This results in a new group containing five ND images (see Fig~\ref{fig:ukbench}(II)). We use Torchvision to generate the new dataset.

\end{itemize}

To evaluate the performance of the hashing-based image deduplication algorithm, we conduct experiments on the \modif{CIFAR10, CIFAR100, and SVHN, respectively. For each dataset, the metrics of hashing time, memory used, and duplicates found for different algorithms are compared under various threshold settings. UKBench and Aug-UKBench are used to evaluate the deduplication performance under different thresholds.}

In the experiments of CNN training and classification, the CIFAR10, CIFAR100, and SVHN datasets after the deduplication are used. For each dataset, the performance of different algorithms are compared from the aspects of training time and validation accuracy.




\subsection{CNN Training}
The ResNet50~\cite{he2016deep} architecture is employed to train image classifications on the deduplicated training sets. CNNs are trained to classify images from 10 classes in the CIFAR10 datasets and 100 classes in the CIFAR100 dataset. For SVHN, the models are trained to classify 10 digits from 0 to 9. We adopt the official train and validation split from Torchvision Datasets~\cite{torchvision}.

During the training process, images are resized to $32 \times 32$ pixels to be compatible with the original dimensions. The parameters of the CNNs are trained using backpropagation with cross-entropy loss. We use Adam optimizer with betas 0.9 and 0.999, respectively, a weight decay of 0, and epsilon of $10^{-8}$. In each training epoch, a mini-batch of 64 images is configured. The initial learning rate is set to 0.01, with a learning rate decay of 10 every 40 epochs. A total of 100 epochs are used to train each model.

The implementation is based on Pytorch~\cite{NEURIPS2019_9015}. We use servers with hardware configured with CPU Intel i9-9900K clocked at 3.60GHz, 32G ddr4 RAM, and Nvidia RTX 2080Ti GPU. For each deduplicated dataset, we train five ResNet50 with each using a different initialization seed.

\section{Evaluation Results}\label{sec:result}

\subsection{Image Deduplication Performance}
\begin{figure}[t!]
\centering
\setlength{\tabcolsep}{5pt} 
\begin{tabular}{c}
\includegraphics[clip,width=.35\textwidth]{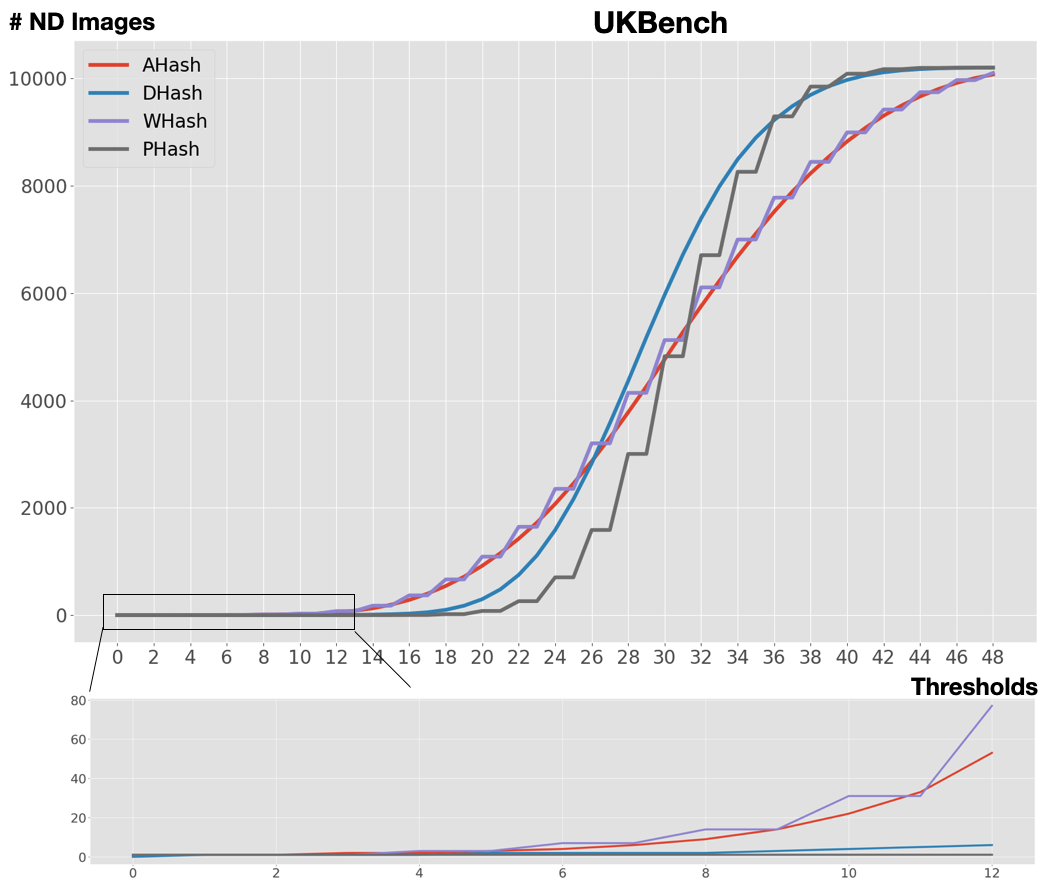}
\end{tabular}
\caption{Number of median ND images found in UKBench dataset when threshold ranges from 0 to 50.}
\label{fig:thres_ukbench}
\end{figure}

\begin{figure}[t!]
\centering
\setlength{\tabcolsep}{5pt} 
\begin{tabular}{c}
\includegraphics[clip,width=.35\textwidth]{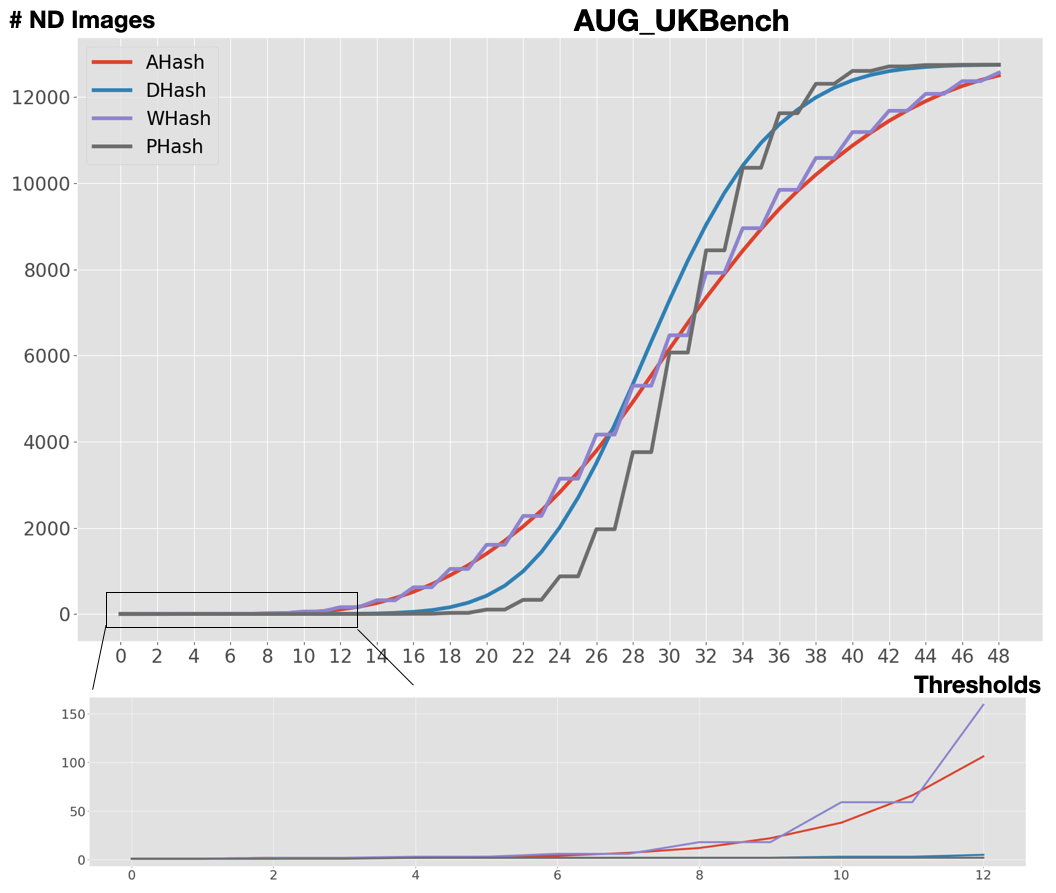}
\end{tabular}
\caption{Number of median ND images found in AUG\_UKBench dataset when threshold ranges from 0 to 50.}
\label{fig:thres_aug_ukbench}
\end{figure}

\modif{We start by evaluating the deduplication efficacy under different thresholds. We plot the median ND images found for each hashing algorithm when threshold varies from 0 to 50. As shown in Fig~\ref{fig:thres_ukbench},\ref{fig:thres_aug_ukbench}, when the threshold is above 12, found ND images are over 150 for AUG\_UKBench. Given the property of UKBench and AUG\_UKBench, ND images should be 4-5, indicating an optimal threshold be 6 to 8. For an in-depth comparison, we use thredholds from 0 to 12 for the rest of the experiments.}

\modif{Next}, We evaluate the performance of different deduplication algorithms presented in Section \ref{subsec:hashing}. For each hashing algorithm, we record the hashing time, the memory cost, and the mean NDs for each image. In detail, $Time_{Hash}$ is the mean time with standard deviation in seconds, $Mem$ is the mean memories with standard deviations in megabytes, and $Dup$ is the mean number of ND images detected for each input.



\begin{figure}[t!]
\centering
\setlength{\tabcolsep}{5pt} 
\begin{tabular}{cc}
\includegraphics[trim=15 10 15 10,clip,width=.1\textwidth]{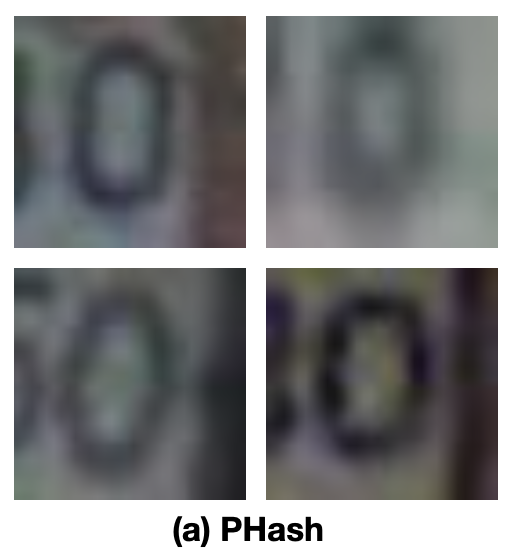} &
\includegraphics[trim=15 10 15 10,clip,width=.1\textwidth]{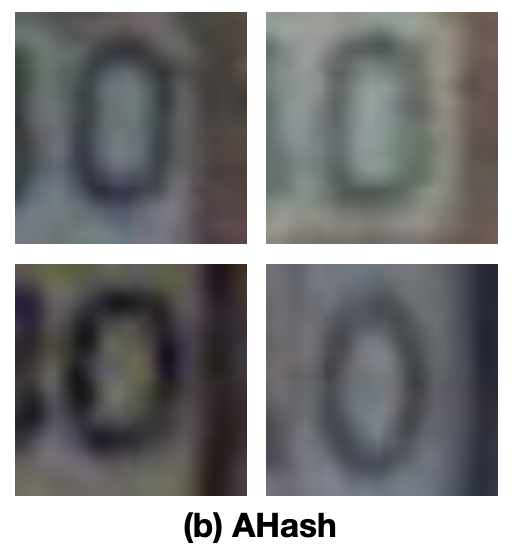} \\
\includegraphics[trim=15 10 15 10,clip,width=.1\textwidth]{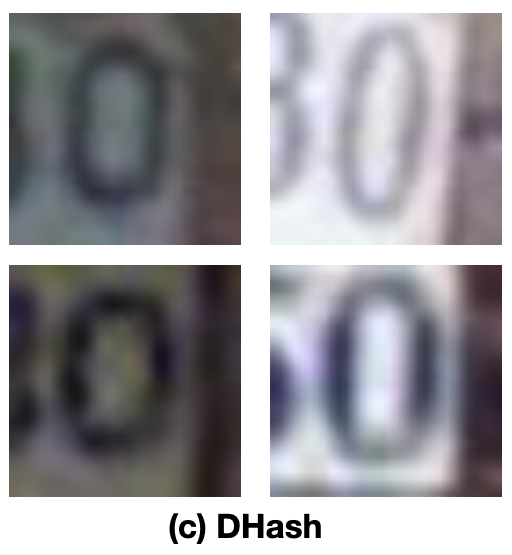} 
&
\includegraphics[trim=15 10 15 10,clip,width=.1\textwidth]{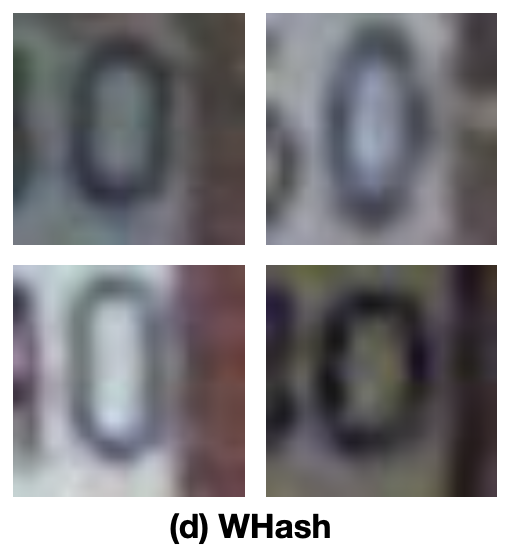} 
\end{tabular}

\caption{Duplicates detected by different hashing algorithms when $th$ is 10. The top left image in (a)-(d) is the same image for better comparison.}
\label{fig:dup_found}
\end{figure}

In the CIFAR10 dataset (Table~\ref{tab:cifar10}), when $th$ increases from 0 to 12, PHash and DHash detect duplicates of 1 to 1.64 and 1.90, respectively. When setting $th$ to 12, AHash and WHash yield candidates of up to 34 and 38.32, respectively. Due to the similar data distribution for CIFAR100, we obtain comparable results. The result for SVHN is in Table~\ref{tab:svhn}. There are generally more duplicates in the datasets. When setting $th$ to 12, PHash detects 3.79 duplicates, and DHash, AHash, WHash detects duplicates of 26.83, 52.32, and 57.30, respectively. In Fig~\ref{fig:dup_found}, we plot four exemplar duplicates detected by each hashing algorithm in the SVHN dataset with $th$ setting to 10. The top left image in (a) - (d) is the same for a fair comparison. It shows that PHash and AHash are more effective in detecting duplicates with similar visual displays. This confirms our hypothesis that SVHN dataset contains many NDs. 

In summary, PHash and DHash achieve the most stable performance under various threshold settings. For the hashing time, PHash, DHash, and AHash prove to be more efficient than WHash. However, WHash detects a significantly escalated number of duplicates when the threshold increases. Further, to achieve a larger deduplication ratio under the same threshold, WHash and AHash are better candidates by detecting more ND images.

\begin{table}[ht!]
 \centering
 \caption{Deduplication Results on the CIFAR10 dataset.} 
\label{tab:cifar10}
\fontsize{7}{8.5}\selectfont
\begin{tabular}{ |lcccc|  }
 \hline
 Method   & $th$ & $Time_{Hash}$ & $Mem$ & $Dup$ \\
 \hline
\multirow{7}{*}{PHash} & 0 & 2.35 \ppm 0.24 & 12.13 \ppm 11.80 & 1.00 \ppm 0.00 \\
& 2 & 2.14 \ppm 0.01 & 11.30 \ppm 12.41 & 1.01 \ppm 0.17 \\
& 4 & 2.17 \ppm 0.01 & 13.95 \ppm 10.36 & 1.12 \ppm 0.60 \\
& 6 & 2.20 \ppm 0.05 & 17.18 \ppm 8.39 & 1.17 \ppm 0.80 \\
& 8 & 2.21 \ppm 0.05 & 12.19 \ppm 11.92 & 1.25 \ppm 0.94 \\
& 10 & 2.21 \ppm 0.04 & 15.92 \ppm 8.90 & 1.33 \ppm 1.08 \\
& 12 & 2.20 \ppm 0.05 & 15.56 \ppm 9.30 & 1.64 \ppm 1.55 \\
\midrule
\multirow{7}{*}{AHash} & 0 & 1.92 \ppm 0.07 & 19.99 \ppm 6.09 & 2.41 \ppm 5.95 \\
& 2 & 1.86 \ppm 0.00 & 14.39 \ppm 10.21 & 5.29 \ppm 15.17 \\
& 4 & 1.90 \ppm 0.07 & 14.64 \ppm 9.82 & 6.63 \ppm 23.67 \\
& 6 & 1.90 \ppm 0.06 & 12.80 \ppm 11.44 & 8.33 \ppm 34.42 \\
& 8 & 1.95 \ppm 0.07 & 12.24 \ppm 11.93 & 11.71 \ppm 48.71 \\
& 10 & 1.98 \ppm 0.08 & 12.38 \ppm 11.75 & 19.38 \ppm 75.04 \\
& 12 & 1.94 \ppm 0.09 & 12.12 \ppm 12.05 & 34.00 \ppm 116.17 \\
\midrule
\multirow{7}{*}{DHash} & 0 & 1.94 \ppm 0.05 & 14.05 \ppm 10.45 & 1.00 \ppm 0.00 \\
& 2 & 1.90 \ppm 0.05 & 12.14 \ppm 11.81 & 1.03 \ppm 0.18 \\
& 4 & 1.94 \ppm 0.04 & 11.82 \ppm 12.01 & 1.10 \ppm 0.37 \\
& 6 & 1.90 \ppm 0.05 & 11.95 \ppm 12.08 & 1.21 \ppm 0.81 \\
& 8 & 1.88 \ppm 0.01 & 19.33 \ppm 6.73 & 1.35 \ppm 1.31 \\
& 10 & 1.90 \ppm 0.05 & 13.60 \ppm 10.83 & 1.58 \ppm 1.89 \\
& 12 & 1.95 \ppm 0.12 & 14.81 \ppm 9.83 & 1.90 \ppm 2.80 \\
\midrule
\multirow{7}{*}{WHash} & 0 & 15.87 \ppm 0.31 & 11.68 \ppm 12.00 & 2.49 \ppm 13.30 \\
& 2 & 17.02 \ppm 0.62 & 12.98 \ppm 11.13 & 6.06 \ppm 29.61 \\
& 4 & 17.49 \ppm 0.84 & 14.05 \ppm 10.26 & 7.59 \ppm 35.91 \\
& 6 & 17.37 \ppm 0.70 & 12.13 \ppm 11.89 & 9.11 \ppm 41.18 \\
& 8 & 17.36 \ppm 0.74 & 10.89 \ppm 12.72 & 13.32 \ppm 54.44 \\
& 10 & 17.02 \ppm 0.59 & 11.46 \ppm 12.36 & 21.68 \ppm 83.83 \\
& 12 & 17.66 \ppm 0.90 & 10.61 \ppm 13.31 & 38.32 \ppm 131.82 \\
 \hline
\end{tabular}
\end{table}

\begin{table}[ht!]
 \centering
 \caption{Deduplication Results on the SVHN dataset.} 
\label{tab:svhn}
\fontsize{7}{8.5}\selectfont
\begin{tabular}{ |lcccc|  }
 \hline
 Method   & $th$ & $Time_{Hash}$ & $Mem$ & $Dup$ \\
 \hline
\multirow{7}{*}{PHash} & 0 & 3.54 \ppm 0.09 & 23.48 \ppm 8.02 & 1.00 \ppm 0.00 \\
& 2 & 3.60 \ppm 0.05 & 26.01 \ppm 5.73 & 1.00 \ppm 0.00 \\
& 4 & 3.60 \ppm 0.03 & 24.60 \ppm 7.06 & 1.03 \ppm 0.23 \\
& 6 & 3.59 \ppm 0.13 & 21.76 \ppm 8.79 & 1.14 \ppm 0.41 \\
& 8 & 3.52 \ppm 0.09 & 30.06 \ppm 3.83 & 1.42 \ppm 1.03 \\
& 10 & 3.57 \ppm 0.04 & 33.03 \ppm 0.83 & 2.07 \ppm 2.21 \\
& 12 & 3.54 \ppm 0.09 & 26.17 \ppm 6.36 & 3.79 \ppm 5.01 \\
\midrule
\multirow{7}{*}{AHash} & 0 & 3.20 \ppm 0.02 & 29.59 \ppm 4.44 & 1.98 \ppm 4.25 \\
& 2 & 3.21 \ppm 0.04 & 26.07 \ppm 6.66 & 2.93 \ppm 6.65 \\
& 4 & 3.21 \ppm 0.05 & 27.03 \ppm 8.36 & 4.42 \ppm 10.56 \\
& 6 & 3.16 \ppm 0.04 & 26.61 \ppm 4.74 & 7.60 \ppm 17.80 \\
& 8 & 3.26 \ppm 0.06 & 20.28 \ppm 8.99 & 13.71 \ppm 33.07 \\
& 10 & 3.28 \ppm 0.03 & 20.05 \ppm 9.21 & 25.87 \ppm 64.77 \\
& 12 & 3.40 \ppm 0.07 & 19.47 \ppm 9.55 & 52.32 \ppm 127.49 \\
\midrule
\multirow{7}{*}{DHash} & 0 & 3.08 \ppm 0.13 & 26.97 \ppm 5.59 & 1.85 \ppm 3.80 \\
& 2 & 3.12 \ppm 0.00 & 26.87 \ppm 5.47 & 3.08 \ppm 7.88 \\
& 4 & 3.04 \ppm 0.09 & 27.07 \ppm 5.97 & 3.70 \ppm 10.16 \\
& 6 & 3.09 \ppm 0.03 & 23.28 \ppm 8.53 & 5.08 \ppm 14.61 \\
& 8 & 3.04 \ppm 0.09 & 22.85 \ppm 7.82 & 8.01 \ppm 22.42 \\
& 10 & 3.13 \ppm 0.01 & 19.46 \ppm 9.60 & 14.14 \ppm 38.57 \\
& 12 & 3.21 \ppm 0.02 & 20.33 \ppm 9.05 & 26.83 \ppm 70.01 \\
\midrule
\multirow{7}{*}{WHash} & 0 & 26.99 \ppm 0.55 & 27.98 \ppm 4.97 & 1.82 \ppm 3.24 \\
& 2 & 27.07 \ppm 0.78 & 32.15 \ppm 0.89 & 3.00 \ppm 5.27 \\
& 4 & 29.75 \ppm 1.47 & 22.12 \ppm 7.93 & 4.94 \ppm 11.12 \\
& 6 & 29.70 \ppm 1.54 & 19.26 \ppm 9.30 & 8.37 \ppm 22.25 \\
& 8 & 29.82 \ppm 1.66 & 19.73 \ppm 9.05 & 14.41 \ppm 41.58 \\
& 10 & 29.82 \ppm 1.44 & 19.33 \ppm 9.23 & 27.62 \ppm 77.93 \\
& 12 & 29.75 \ppm 1.41 & 19.24 \ppm 9.37 & 57.30 \ppm 145.76 \\
 \hline
\end{tabular}
\end{table}

\begin{table}[ht!]
 \centering
 \caption{CNN Results on the deduplicated CIFAR10 dataset.}
\label{tab:cifar10_training}
\fontsize{6}{8}\selectfont
\begin{tabular}{ |lccccc|  }
 \hline
 Method & Threshold & $D_{train}$ & $D_{val}$ & $Acc$ & $Time_{train}$ \\
 \hline
\multirow{7}{*}{PHash} & 0 & 49988 & 9995 & 0.81 \ppm 0.00 & 9028.62 \ppm 23.07 \\
& 2 & 49888 & 9976 & 0.81 \ppm 0.00 & 9059.99 \ppm 8.73 \\
& 4 & 49709 & 9940 & 0.81 \ppm 0.01 & 9034.37 \ppm 12.68 \\
& 6 & 49475 & 9889 & 0.80 \ppm 0.01 & 8978.31 \ppm 14.99 \\
& 8 & 48991 & 9790 & 0.80 \ppm 0.00 & 7574.20 \ppm 30.28 \\
& 10 & 47361 & 9446 & 0.79 \ppm 0.00 & 6377.94 \ppm 127.48 \\
& 12 & 41233 & 8188 & 0.77 \ppm 0.00 & 3778.55 \ppm 7.87 \\
\midrule
\multirow{7}{*}{AHash} & 0 & 49297 & 9865 & 0.80 \ppm 0.01 & 8949.10 \ppm 12.32 \\
& 2 & 47243 & 9421 & 0.79 \ppm 0.00 & 8598.39 \ppm 6.97 \\
& 4 & 43145 & 8626 & 0.78 \ppm 0.01 & 7857.52 \ppm 1.20 \\
& 6 & 36659 & 7331 & 0.77 \ppm 0.01 & 6672.30 \ppm 7.37 \\
& 8 & 28271 & 5713 & 0.76 \ppm 0.01 & 5163.91 \ppm 4.71 \\
& 10 & 19415 & 3959 & 0.71 \ppm 0.02 & 3547.27 \ppm 7.38 \\
& 12 & 11754 & 2411 & 0.65 \ppm 0.01 & 1908.50 \ppm 1.55 \\
\midrule
\multirow{7}{*}{DHash} & 0 & 49993 & 9997 & 0.81 \ppm 0.01 & 9074.42 \ppm 21.06 \\
& 2 & 49927 & 9980 & 0.80 \ppm 0.01 & 9087.82 \ppm 13.81 \\
& 4 & 49748 & 9947 & 0.81 \ppm 0.00 & 9043.53 \ppm 4.01 \\
& 6 & 49500 & 9901 & 0.80 \ppm 0.00 & 8982.48 \ppm 14.75 \\
& 8 & 49024 & 9822 & 0.79 \ppm 0.01 & 7519.38 \ppm 17.74 \\
& 10 & 47604 & 9562 & 0.81 \ppm 0.00 & 4041.21 \ppm 16.33 \\
& 12 & 42324 & 8528 & 0.79 \ppm 0.00 & 1869.77 \ppm 2.51 \\
\midrule
\multirow{7}{*}{WHash} & 0 & 49133 & 9828 & 0.81 \ppm 0.00 & 8925.77 \ppm 11.83 \\
& 2 & 46824 & 9376 & 0.81 \ppm 0.00 & 8520.37 \ppm 8.03 \\
& 4 & 42268 & 8445 & 0.79 \ppm 0.01 & 7701.92 \ppm 7.23 \\
& 6 & 35243 & 7051 & 0.77 \ppm 0.00 & 6425.45 \ppm 7.93 \\
& 8 & 26702 & 5353 & 0.75 \ppm 0.02 & 4888.08 \ppm 7.30 \\
& 10 & 17960 & 3614 & 0.71 \ppm 0.01 & 3304.23 \ppm 4.92 \\
& 12 & 10506 & 2133 & 0.57 \ppm 0.05 & 1983.27 \ppm 8.48 \\
\midrule
\multicolumn{2}{|l}{Original} & 50000 & 10000 & 0.81 \ppm 0.00 & 9276.00 \ppm 6.96 \\
\hline
\end{tabular}
\end{table}

\begin{table}[ht!]
 \centering
 \caption{CNN Results on the deduplicated SVHN dataset.}
\label{tab:svhn_training}
\fontsize{6}{8}\selectfont
\begin{tabular}{ |lccccc|  }
 \hline
 Method & Threshold & $D_{train}$ & $D_{val}$ & $Acc$ & $Time_{train}$ \\
 \hline
\multirow{7}{*}{PHash} & 0 & 73223 & 26032 & 0.93 \ppm 0.01 & 14325.53 \ppm 115.62 \\
& 2 & 73195 & 26031 & 0.93 \ppm 0.00 & 14352.74 \ppm 38.40 \\
& 4 & 73047 & 25993 & 0.69 \ppm 0.35 & 14333.61 \ppm 51.98 \\
& 6 & 72246 & 25670 & 0.94 \ppm 0.00 & 13901.18 \ppm 99.00 \\
& 8 & 68861 & 24258 & 0.93 \ppm 0.00 & 8319.84 \ppm 54.96 \\
& 10 & 59584 & 20442 & 0.92 \ppm 0.00 & 3679.52 \ppm 240.56 \\
& 12 & 42248 & 13977 & 0.38 \ppm 0.23 & 2354.82 \ppm 32.24 \\
\midrule
\multirow{7}{*}{AHash} & 0 & 72475 & 25863 & 0.93 \ppm 0.00 & 14173.13 \ppm 124.37 \\
& 2 & 68820 & 24201 & 0.94 \ppm 0.00 & 13523.15 \ppm 38.63 \\
& 4 & 60219 & 19904 & 0.93 \ppm 0.00 & 11815.40 \ppm 13.90 \\
& 6 & 46980 & 14512 & 0.88 \ppm 0.03 & 9181.83 \ppm 29.28 \\
& 8 & 32774 & 9487 & 0.62 \ppm 0.22 & 6326.06 \ppm 84.77 \\
& 10 & 20458 & 5679 & 0.75 \ppm 0.03 & 3983.87 \ppm 44.33 \\
& 12 & 11331 & 3083 & 0.82 \ppm 0.00 & 2007.78 \ppm 9.04 \\
\midrule
\multirow{7}{*}{DHash} & 0 & 72976 & 25986 & 0.92 \ppm 0.01 & 14291.59 \ppm 109.03 \\
& 2 & 71427 & 25667 & 0.92 \ppm 0.01 & 14067.99 \ppm 17.76 \\
& 4 & 66601 & 24549 & 0.93 \ppm 0.00 & 13148.18 \ppm 7.88 \\
& 6 & 56831 & 21836 & 0.92 \ppm 0.02 & 11185.53 \ppm 39.77 \\
& 8 & 43470 & 17476 & 0.91 \ppm 0.01 & 8144.05 \ppm 19.04 \\
& 10 & 29215 & 12203 & 0.86 \ppm 0.03 & 3107.53 \ppm 17.87 \\
& 12 & 17217 & 7374 & 0.87 \ppm 0.01 & 2559.07 \ppm 11.07 \\
\midrule
\multirow{7}{*}{WHash} & 0 & 72427 & 25633 & 0.93 \ppm 0.00 & 14129.86 \ppm 111.01 \\
& 2 & 68470 & 23432 & 0.94 \ppm 0.00 & 13432.39 \ppm 63.33 \\
& 4 & 59320 & 19056 & 0.93 \ppm 0.01 & 11619.81 \ppm 24.09 \\
& 6 & 45972 & 14021 & 0.67 \ppm 0.35 & 8959.56 \ppm 30.83 \\
& 8 & 31498 & 9195 & 0.57 \ppm 0.30 & 6106.20 \ppm 9.55 \\
& 10 & 18955 & 5468 & 0.86 \ppm 0.01 & 3752.69 \ppm 23.90 \\
& 12 & 10270 & 2910 & 0.16 \ppm 0.01 & 2058.67 \ppm 10.99 \\
\midrule
\multicolumn{2}{|l}{Original} & 73255 & 26032 & 0.92 \ppm 0.02 & 14269.56 \ppm 18.37 \\
 \hline
\end{tabular}
\end{table}

\subsection{CNN Training Performance}
We study the effectiveness of deduplicated datasets in the CNN training application. We train CNNs on the deduplicated CIFAR10, CIFAR100, and SVHN datasets. Training and validation split is from the official Torchvision Datasets~\cite{torchvision}. The number of images after the deduplication in the training dataset $D_{train}$ and in the validation dataset $D_{val}$ are reported, respectively. On each training dataset, we train ResNet50~\cite{he2016deep} on 5 random initializations. The mean and standard deviation of validation accuracy $Acc$ and training time $Time_{train}$ are calculated. The performance in the original un-processed dataset is also presented.

Table~\ref{tab:cifar10_training} shows the results for the deduplicated CIFAR10 dataset. PHash and DHash demonstrate a consistent validation accuracy of 0.81 when using a threshold up to 4. The image number of the training dataset is reduced to 49709, and the validation data is slightly down to 9940. We observe a speedup of 200 seconds in training time compared with the original dataset. AHash and WHash algorithms exhibit a small accuracy drop but detect more duplicates overall. Specifically, after the deduplication with a threshold of 12, PHash and DHash have over 40000 images left in the training set, whereas AHash and WHash have less than 12000 images.

Similar results for deduplicated CIFAR100 dataset are obtained. PHash and DHash maintain a consistent accuracy from 0.43 to 0.40 when the threshold increases from 0 to 12. And the dataset size ranges from over 49900 to 42969 and 38577, respectively. In contrast, AHash and WHash have the largest fluctuation with accuracy drops from 0.43 to 0.12 lowest, and dataset size varies from about 49000 to over 10000. 

Table~\ref{tab:svhn_training} illustrates the results for the deduplicated SVHN datasets. We can see PHash achieves the best overall accuracy. Validation accuracy maintains 0.92 even when the threshold is set to 10. The dataset is reduced to 59584 images, and the training time is reduced to only 25\%. AHash, DHash, and WHash get comparable results on the dataset size and training time. But WHash has the overall lowest accuracy, especially when the threshold is set to 12. 

In summary, \modif{as suggested from Fig~\ref{fig:thres_ukbench},\ref{fig:thres_aug_ukbench}, a proper deduplication threshold for safely removing ND images should be set between 6 to 8. Futher experiments from the CNN-based perspective suggest that}
PHash and DHash again retain relatively stable accuracy when the threshold increase from 0 to 6. The deduplicated dataset is reduced by 77\% largest, with a 30\% speedup in training time. \modif{We therefore conclude that using PHash and DHash with the threshold set to 6 would obtain reduced training time, and storage cost, while maintaining the true diversity of the datasets. As only the ND images are removed, and validation accuracy is retained. In scenarios where we want sacrifice some accuracy in trade of shortened training time, such as neural architecture search, or hyper-parameter optimization, we would suggest a threshold of 10 to obtain large dataset reduction (reduced by 77\% for SVHN, with the accuracy dropped by 6\%.
}

\section{Related Works}\label{sec:related}
Image hashing algorithms are proposed by compressing the original image into unique hash sequences~\cite{pan2005fast}. In the domain of image deduplication, such image hashing algorithms yield identical hashes for images under slight transformation such as rotation, resizing, or color balancing, etc. 

Few applications are designed to meet above constraints based on image compression algorithms using Fourier transformation~\cite{swaminathan2006robust}. Venkatesan et al.~\cite{venkatesan2000robust} associate image hashes using coefficients of Discrete Wavelet Transform (DWT). In another approach, Discrete Cosine Transform (DCT)~\cite{ahmed1974discrete} is adopted to generate the hash sequence. Lin et al.~\cite{lin2001robust} extend the work by associating invariant information between DCT coefficients. Layek et al.~\cite{layek2016fast} leverage Locality Sensitive Hashing (LSH) to identify similarity for images in social media. Zhou et al.~\cite{zhou2017fast} propose applying LSH on image blocks to detect ND images for visual sensor networks. Rashid et al.~\cite{rashid2016secure} propose Set Partitioning in Hierarchical Trees(SPIHT) image compression in combination with partial encryption to generate a hash for the image. In contrast to these works, this paper focuses on quantifying the efficacy of hashing algorithms from a CNN training perspective.

\section{Discussion and Conclusion}\label{sec:conclu}

This paper is motivated by the finding that near-duplicate images contribute little to no accuracy improvement in CNN. Moreover, disk storage, training time, and resources consumption pose a huge obstacle for users. Therefore, we propose a framework named \textit{CE-Dedup} to combine the hashing-based image deduplication with downstream CNN-based image classification tasks. Extensive experiments are conducted to evaluate the performance of both parts.

The results confirm that from the image deduplication perspective, PHash and DHash yield consistent performance in terms of hashing time cost, memory consumption, and duplicates detected when the threshold varies from 0 to 12. The duplicates detected by AHash and Whash increase drastically when the threshold increases. The memory consumption for WHash is consistently high which might be due to the computation of wavelet transformation. 



To further validate the efficacy of image deduplication on neural nets, we aim to evaluate a wider range of image deduplication algorithms, including SIFT~\cite{lowe2004distinctive}, SURF~\cite{bay2006surf}, ORB~\cite{rublee2011orb}, feature extraction-based approaches. As future work, we would also like to test a more diverse set of image datasets such as MS-COCO~\cite{lin2014microsoft} for object detection, and more CNN architectures.

\bibliographystyle{plain}
\bibliography{references}

\begin{thebibliography}{10}

\bibitem{ahmed1974discrete}
Nasir Ahmed, T\_ Natarajan, and Kamisetty~R Rao.
\newblock Discrete cosine transform.
\newblock {\em IEEE transactions on Computers}, 100(1):90--93, 1974.

\bibitem{bay2006surf}
Herbert Bay, Tinne Tuytelaars, and Luc Van~Gool.
\newblock Surf: Speeded up robust features.
\newblock In {\em European conference on computer vision}, pages 404--417.
  Springer, 2006.

\bibitem{bloice2017augmentor}
Marcus~D Bloice, Christof Stocker, and Andreas Holzinger.
\newblock Augmentor: an image augmentation library for machine learning.
\newblock {\em arXiv preprint arXiv:1708.04680}, 2017.

\bibitem{brown2020language}
Tom~B Brown, Benjamin Mann, Nick Ryder, Melanie Subbiah, Jared Kaplan, Prafulla
  Dhariwal, Arvind Neelakantan, Pranav Shyam, Girish Sastry, Amanda Askell,
  et~al.
\newblock Language models are few-shot learners.
\newblock {\em arXiv preprint arXiv:2005.14165}, 2020.

\bibitem{chamoso2017hash}
Pablo Chamoso, Alberto Rivas, Javier~J Mart{\'\i}n-Limorti, and Sara
  Rodr{\'\i}guez.
\newblock A hash based image matching algorithm for social networks.
\newblock In {\em International Conference on Practical Applications of Agents
  and Multi-Agent Systems}, pages 183--190. Springer, 2017.

\bibitem{deng2009imagenet}
Jia Deng, Wei Dong, Richard Socher, Li-Jia Li, Kai Li, and Li~Fei-Fei.
\newblock Imagenet: A large-scale hierarchical image database.
\newblock In {\em 2009 IEEE conference on computer vision and pattern
  recognition}, pages 248--255. Ieee, 2009.

\bibitem{doi_ulrich_2021}
Kenji Doi and Sam Ulrich.
\newblock knjcode/imgdupes.
\newblock url={https://github.com/knjcode/imgdupes}, 2021.

\bibitem{goyal2017accurate}
Priya Goyal, Piotr Doll{\'a}r, Ross Girshick, Pieter Noordhuis, Lukasz
  Wesolowski, Aapo Kyrola, Andrew Tulloch, Yangqing Jia, and Kaiming He.
\newblock Accurate, large minibatch sgd: Training imagenet in 1 hour.
\newblock {\em arXiv preprint arXiv:1706.02677}, 2017.

\bibitem{he2016deep}
Kaiming He, Xiangyu Zhang, Shaoqing Ren, and Jian Sun.
\newblock Deep residual learning for image recognition.
\newblock In {\em Proceedings of the IEEE conference on computer vision and
  pattern recognition}, pages 770--778, 2016.

\bibitem{idealods2019imagededup}
Tanuj Jain, Christopher Lennan, Zubin John, and Dat Tran.
\newblock Imagededup.
\newblock url={https://github.com/idealo/imagededup}, 2019.

\bibitem{cifar100}
Alex Krizhevsky, Vinod Nair, and Geoffrey Hinton.
\newblock Cifar-100 (canadian institute for advanced research).

\bibitem{layek2016fast}
Ashish~Kumar Layek, Akash Gupta, Saptarshi Ghosh, and Sekhar Mandal.
\newblock Fast near-duplicate detection from image streams on online social
  media during disaster events.
\newblock In {\em 2016 IEEE Annual India Conference (INDICON)}, pages 1--6.
  IEEE, 2016.

\bibitem{li2020intelligent}
Yibin Li, Yan Song, Lei Jia, Shengyao Gao, Qiqiang Li, and Meikang Qiu.
\newblock Intelligent fault diagnosis by fusing domain adversarial training and
  maximum mean discrepancy via ensemble learning.
\newblock {\em IEEE Transactions on Industrial Informatics}, 17(4):2833--2841,
  2020.

\bibitem{lin2001robust}
Ching-Yung Lin and Shih-Fu Chang.
\newblock A robust image authentication method distinguishing jpeg compression
  from malicious manipulation.
\newblock {\em IEEE Transactions on Circuits and Systems for Video Technology},
  11(2):153--168, 2001.

\bibitem{lin2014microsoft}
Tsung-Yi Lin, Michael Maire, Serge Belongie, James Hays, Pietro Perona, Deva
  Ramanan, Piotr Doll{\'a}r, and C~Lawrence Zitnick.
\newblock Microsoft coco: Common objects in context.
\newblock In {\em European conference on computer vision}, pages 740--755.
  Springer, 2014.

\bibitem{lowe2004distinctive}
David~G Lowe.
\newblock Distinctive image features from scale-invariant keypoints.
\newblock {\em International journal of computer vision}, 60(2):91--110, 2004.

\bibitem{o2015introduction}
Keiron O'Shea and Ryan Nash.
\newblock An introduction to convolutional neural networks.
\newblock {\em arXiv preprint arXiv:1511.08458}, 2015.

\bibitem{pan2005fast}
Feng Pan, Xiao Lin, Susanto Rahardja, Keng~Pang Lim, ZG~Li, Dajun Wu, and
  Si~Wu.
\newblock Fast mode decision algorithm for intraprediction in h. 264/avc video
  coding.
\newblock {\em IEEE Transactions on Circuits and Systems for Video Technology},
  15(7):813--822, 2005.

\bibitem{NEURIPS2019_9015}
Adam Paszke, Sam Gross, Francisco Massa, Adam Lerer, James Bradbury, Gregory
  Chanan, Trevor Killeen, Zeming Lin, Natalia Gimelshein, Luca Antiga, Alban
  Desmaison, Andreas Kopf, Edward Yang, Zachary DeVito, Martin Raison, Alykhan
  Tejani, Sasank Chilamkurthy, Benoit Steiner, Lu~Fang, Junjie Bai, and Soumith
  Chintala.
\newblock Pytorch: An imperative style, high-performance deep learning library.
\newblock In H.~Wallach, H.~Larochelle, A.~Beygelzimer, F.~d\textquotesingle
  Alch\'{e}-Buc, E.~Fox, and R.~Garnett, editors, {\em Advances in Neural
  Information Processing Systems 32}, pages 8024--8035. Curran Associates,
  Inc., 2019.

\bibitem{pytorch2021}
Pytorch.
\newblock pytorch2021.
\newblock url={https://pytorch.org/vision/0.8/datasets.html}, 2021.

\bibitem{torchvision}
Torchvision Pytorch.
\newblock Torchvision dataset.
\newblock url={https://pytorch.org/vision/0.8/datasets.html}, 2021.

\bibitem{rashid2016secure}
Fatema Rashid, Ali Miri, and Isaac Woungang.
\newblock Secure image deduplication through image compression.
\newblock {\em Journal of Information Security and Applications}, 27:54--64,
  2016.

\bibitem{rublee2011orb}
Ethan Rublee, Vincent Rabaud, Kurt Konolige, and Gary Bradski.
\newblock Orb: An efficient alternative to sift or surf.
\newblock In {\em 2011 International conference on computer vision}, pages
  2564--2571. Ieee, 2011.

\bibitem{scott_2020}
Kevin Scott.
\newblock Microsoft teams up with openai to exclusively license gpt-3 language
  model, Sep 2020.

\bibitem{shensa1992discrete}
Mark~J Shensa.
\newblock The discrete wavelet transform: wedding the a trous and mallat
  algorithms.
\newblock {\em IEEE Transactions on signal processing}, 40(10):2464--2482,
  1992.

\bibitem{singh2017robust}
Satendra~Pal Singh and Gaurav Bhatnagar.
\newblock A robust image hashing based on discrete wavelet transform.
\newblock In {\em 2017 IEEE International Conference on Signal and Image
  Processing Applications (ICSIPA)}, pages 440--444. IEEE, 2017.

\bibitem{su2012secure}
Hai Su, Meikang Qiu, and Honggang Wang.
\newblock Secure wireless communication system for smart grid with rechargeable
  electric vehicles.
\newblock {\em IEEE Communications Magazine}, 50(8):62--68, 2012.

\bibitem{swaminathan2006robust}
Ashwin Swaminathan, Yinian Mao, and Min Wu.
\newblock Robust and secure image hashing.
\newblock {\em IEEE Transactions on Information Forensics and security},
  1(2):215--230, 2006.

\bibitem{venkatesan2000robust}
Ramarathnam Venkatesan, S-M Koon, Mariusz~H Jakubowski, and Pierre Moulin.
\newblock Robust image hashing.
\newblock In {\em Proceedings 2000 International Conference on Image Processing
  (Cat. No. 00CH37101)}, volume~3, pages 664--666. IEEE, 2000.

\bibitem{zauner2010implementation}
Christoph Zauner.
\newblock Implementation and benchmarking of perceptual image hash functions.
\newblock 2010.

\bibitem{zhou2017fast}
Zhili Zhou, QM~Jonathan Wu, Fang Huang, and Xingming Sun.
\newblock Fast and accurate near-duplicate image elimination for visual sensor
  networks.
\newblock {\em International Journal of Distributed Sensor Networks},
  13(2):1550147717694172, 2017.

\end{thebibliography}


\end{document}